\documentclass[letterpaper, 10 pt, conference]{ieeeconf}  
\IEEEoverridecommandlockouts                              

\usepackage{comment}
\usepackage[T1]{fontenc}
\usepackage{siunitx}
\usepackage[pdftex]{graphicx}
\usepackage[hyphens]{url}
\usepackage{booktabs}
\usepackage{multirow}
\usepackage{amssymb}
\usepackage{amsmath}
\usepackage[scr]{rsfso}
\usepackage{subfig}
\usepackage{hyperref}

\usepackage{tikz}

\newcommand\copyrighttext{
	\footnotesize 
	\textcopyright~2022 IEEE. Personal use of this material is permitted. Permission from IEEE must be obtained for all other uses, in any current or future media, including reprinting/republishing this material for advertising or promotional purposes, creating new collective works, for resale or redistribution to servers or lists, or reuse of any copyrighted component of this work in other works. DOI: \href{https://doi.org/10.1109/IV51971.2022.9827081}{10.1109/IV51971.2022.9827081}%
	}
\newcommand\copyrightnotice{%
    \begin{tikzpicture}[remember picture,overlay]%
 	\node[anchor=south, xshift=0pt, yshift=4pt] at (current page.south)%
 	{\fbox{\parbox{\dimexpr\textwidth-\fboxsep-\fboxrule\relax}{\copyrighttext}}};%
 	\end{tikzpicture}%
}%
\title{\LARGE \bf
Situation-Aware Environment Perception for Decentralized Automation Architectures
}%
\author{Matti Henning$^1$, Michael Buchholz$^1$ and Klaus Dietmayer$^1$
\thanks{$^{1}$M. Henning, M. Buchholz, and K. Dietmayer are with the Institute of Measurement, Control and Microtechnology at Ulm University, 89081, Ulm, Germany. E-Mail: <firstname>.<lastname>@uni-ulm.de}%
\thanks{This research is accomplished within the UNICAR\emph{agil} project (FKZ
16EMO0290). We acknowledge the financial support for the project by the
German Federal Ministry of Education and Research (BMBF).}
}

\begin{document}

\maketitle%
\copyrightnotice%
\thispagestyle{empty}
\pagestyle{empty}
%
\begin{abstract}
Advances in the field of environment perception for automated agents have resulted in an ongoing increase in generated sensor data.
The available computational resources to process these data are bound to become insufficient for real-time applications.
Reducing the amount of data to be processed by identifying the most relevant data based on the agents' situation, often referred to as \textit{situation-awareness}, has gained increasing research interest, and the importance of complementary approaches is expected to increase further in the near future. 
In this work, we extend the applicability range of our recently introduced concept for situation-aware environment perception to the decentralized automation architecture of the UNICAR\emph{agil} project. 
Considering the specific driving capabilities of the vehicle and using real-world data on target hardware in a post-processing manner, we provide an estimate for the daily reduction in power consumption that accumulates to  36.2\,\%.
While achieving these promising results, we additionally show the need to consider scalability in data processing in the design of software modules as well as in the design of functional systems if the benefits of situation-awareness shall be leveraged optimally.%
\end{abstract}%
%
\section{Introduction}
\label{sec:intro}
Environment perception within automated driving has received significant attention within the last decades. 
This attention is accompanied by a similarly intense increase in employed sensors, sensor modalities, sensor redundancies, and computational hardware to process the large amount of generated data for state-of-the-art automated vehicles~\cite{Liu2021}.
This has mainly been possible due to the drastic increase in the availability of sensors and computational power of processing hardware as well as the drastic decrease in their acquisition cost.
However, the amount of generated data paired with the increasing complexity of processing algorithms is already exceeding the computational resources~\cite{Liu2021}.

One approach to counteract the limitations imposed by the computational resources is \textit{situation-awareness} for automated agents, as summarized by Dahn et al.~\cite{Dahn2018}. 
Originating from the field of \textit{active perception}, where actuated sensor platforms are employed to obtain relevant data of the environment, situation-awareness aims to identify which information an agent perceives is relevant for its task execution. 
By this distinction, the amount of data to be processed can be drastically reduced and real-time capabilities of automated agents can be improved. 
Besides, even where the computational resources are sufficient to process all available data, these methods allow using resources efficiently and conserving energy.
A few years ago, Bajcsy et al.~\cite{Bajcsy2018} surveyed the advances in situation-awareness and active perception within automated vehicles and concluded that existing and new approaches would need to receive increased attention to accelerate future progress within the field. 

Various approaches for selective data processing exist, e.g., \cite{Pal2020} based on image saliency or \cite{Nager2021} a-priori knowledge of the environment. 
Recently, we have published a novel, flexible and modular concept for situation-aware environment perception~\cite{henning2022} that allows the integration of such approaches in a formalized way. 
While we previously focused on a centralized processing architecture for environment perception, in this work, we extend the applicability range towards a decentralized processing architecture as introduced by Keilhoff et al.~\cite{keilhoff2019unicaragil}: the disruptive and modular vehicle prototypes for driverless operation within the publicly funded UNICAR\emph{agil} project~\cite{woopen2018, Woopen2020}.

\subsection{Environment Perception in the \emph{UNICAR}agil Vehicles}
\label{sec:intro:unicar}
\begin{figure}[!t]
    \centering
    \includegraphics[width=.5\linewidth]{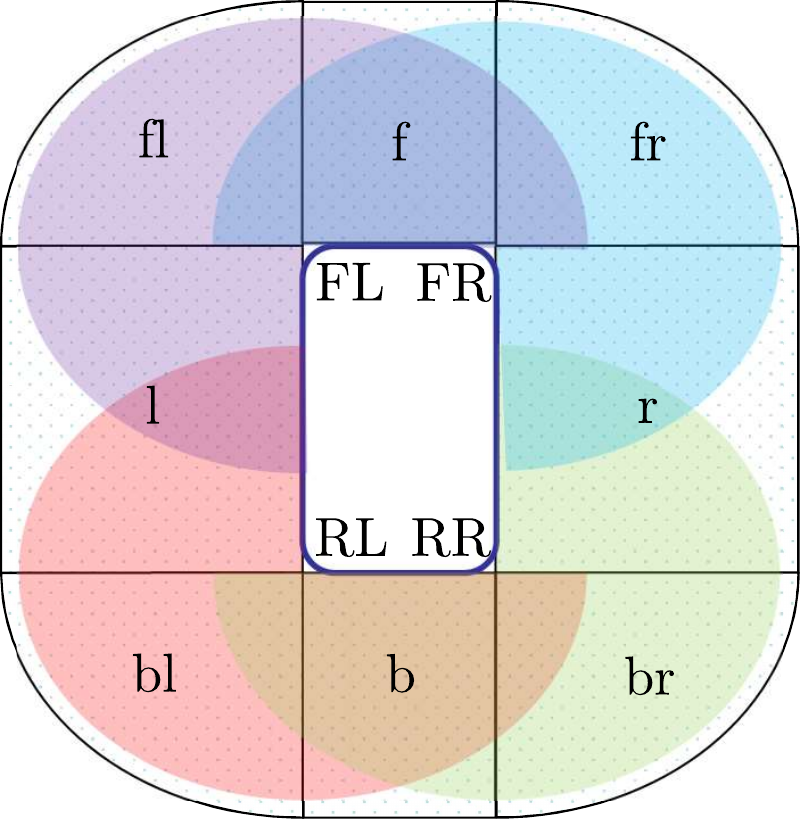}
    \caption{Perception capabilities of an UNICAR\emph{agil} vehicle (blue), including the perception area of the four sensor modules (upper case, filled colors), as well as the separation of the environment into eight regions (lower case, textured).}
    \label{fig:intro:unicar}
\end{figure}
The core concept of the UNICAR\emph{agil} project bases on modularity and service orientation.  Consequently, the environment perception of the prototype vehicles is designed in a modular way, so that all vehicles employ an identical set of four identical \textit{sensor modules}, one at each corner of the vehicle. Each sensor module comprises one lidar sensor, two radar sensors, four cameras, and one inertial sensor, resulting in a multi-modal perception area of \SI[mode=text]{270}{\degree}. A high-end consumer-grade PC accompanies the sensors. 
Fig.~\ref{fig:intro:unicar} illustrates the sensor modules, indicated by either front (F) or rear (R) and either left (L) or right (R) for the respective vehicle corners, as well as their \SI[mode=text]{270}{\degree} perception area. 

The decentralized processing in each sensor module generates an independent environment model with both model-based as well as model-free representations of the surroundings of the vehicle. 
Reflecting the service orientation, the operation mode of the sensor modules can be actively controlled during run-time via the automotive service-oriented architecture  (ASOA)~\cite{mokhtarian2020}.
The independent sensor module representations of the environment are transmitted to a centralized processing unit, the \textit{cerebrum}. Here, after high-level fusion, the interface to the planning modules is realized, which is outside of the scope of this work. A more detailed description of UNICAR\emph{agil} automation concept and the perception system can be found in \cite{Buchholz2020}.

\subsection{Contribution}
In this work, we extend our concept for situation-aware environment perception from~\cite{henning2022} towards decentralized automation architectures with integrated software and hardware modules, i.e., \textit{sensor modules}, and show that also these new, disruptive architectures can be easily considered within this concept.
Besides, instead of  monitoring the hardware load of the system to verify the effectiveness of the application, we measure the actual power consumption of the system and provide estimates on the reduction of consumption within our experiments.
Lastly, with our application results, we show the necessity for considering situation-awareness in the design of software modules and architectures for environment perception to maximize the benefits obtained by situation-awareness.

\section{Situation-Aware Environment Perception}
This section briefly summarizes \textit{awareness processing}~\cite{henning2022}, our recently published concept for situation-aware environment perception.
We refer the interested reader to the publication for more details. Second, we present its application to the decentralized automated vehicle architecture within the UNICAR\emph{agil} project.

\subsection{Awareness Processing}
The concept roots on the assumption that, for any task an automated agent needs to solve, some information about its environment is more relevant than other information. Further, the external circumstances, i.e., the situation of the agent, might influence what information is relevant, even though the task remains identical.

\begin{figure}[!t]
    \centering
    \includegraphics[width=.7\linewidth]{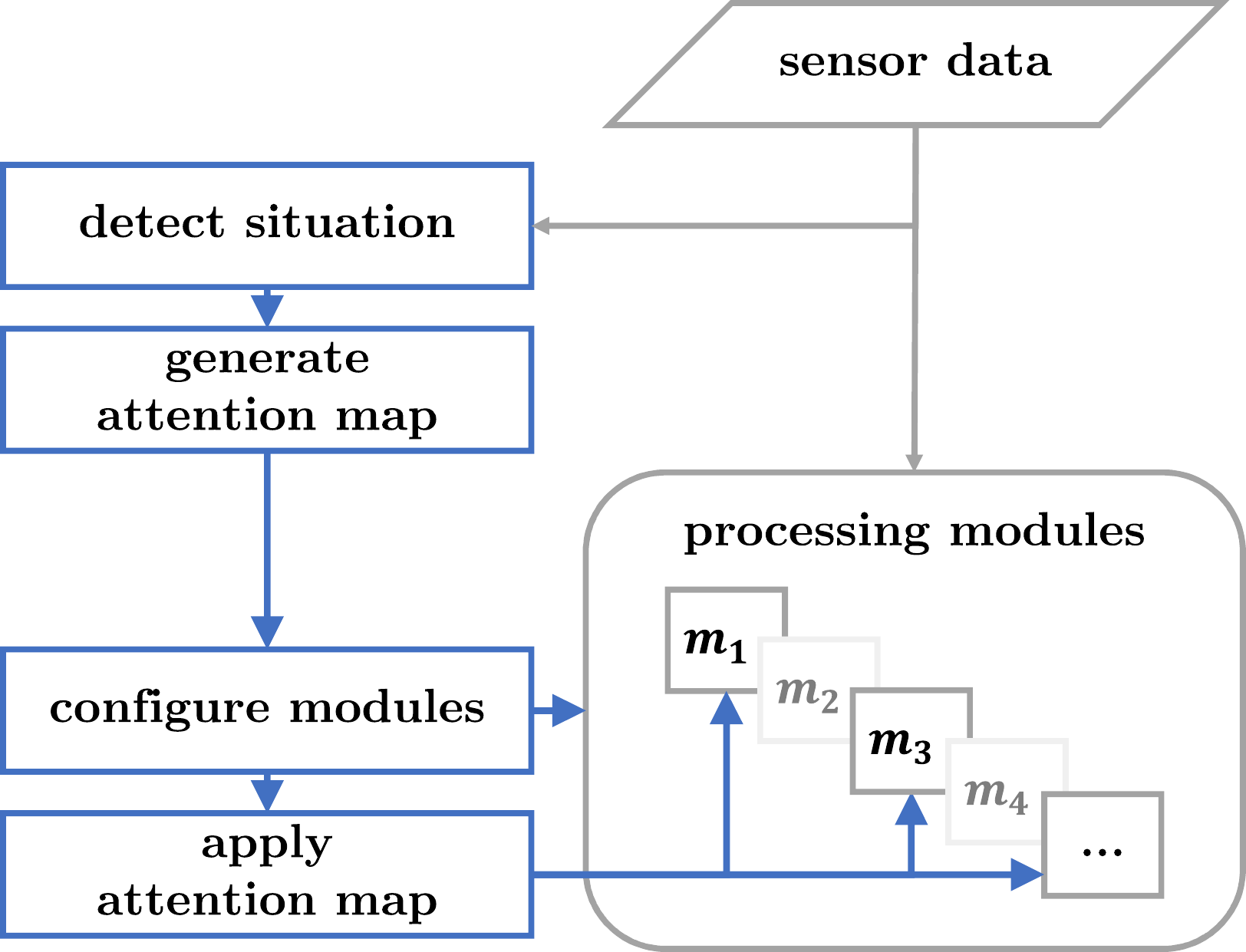}
    \caption{Simplified block diagram of the functional interaction between \textit{awareness processing} (blue) and an existing perception processing chain (gray). Inactive modules are indicated by fading. Diagram adapted from~\cite{henning2022}.}
    \label{fig:saep:block}
\end{figure}

With these assumptions established, the key elements of \textit{awareness processing} are first to describe the relevant regions within the environment and define requirements for the environment perception in these relevant regions. The relevant regions need to be disjoint and depend on the agent's situation. They can be described by any geometric form, e.g., as a grid. The requirements are represented via an attention map corresponding to the regions of the environment. 
Second, the agent's environment perception software is configured to optimally meet the defined requirements of the attention map.
Lastly, the attention map, representing the relevant and non-relevant regions, is fed into the configured perception modules, so that the processing resources can be allocated efficiently. 
Fig.~\ref{fig:saep:block} presents an overview of the key elements and their interaction with an existing processing chain for environment perception. 

\subsubsection{Situation Detection and Attention Map Generation}
\begin{figure}[!t]
    \centering
    \includegraphics[width=.5\linewidth]{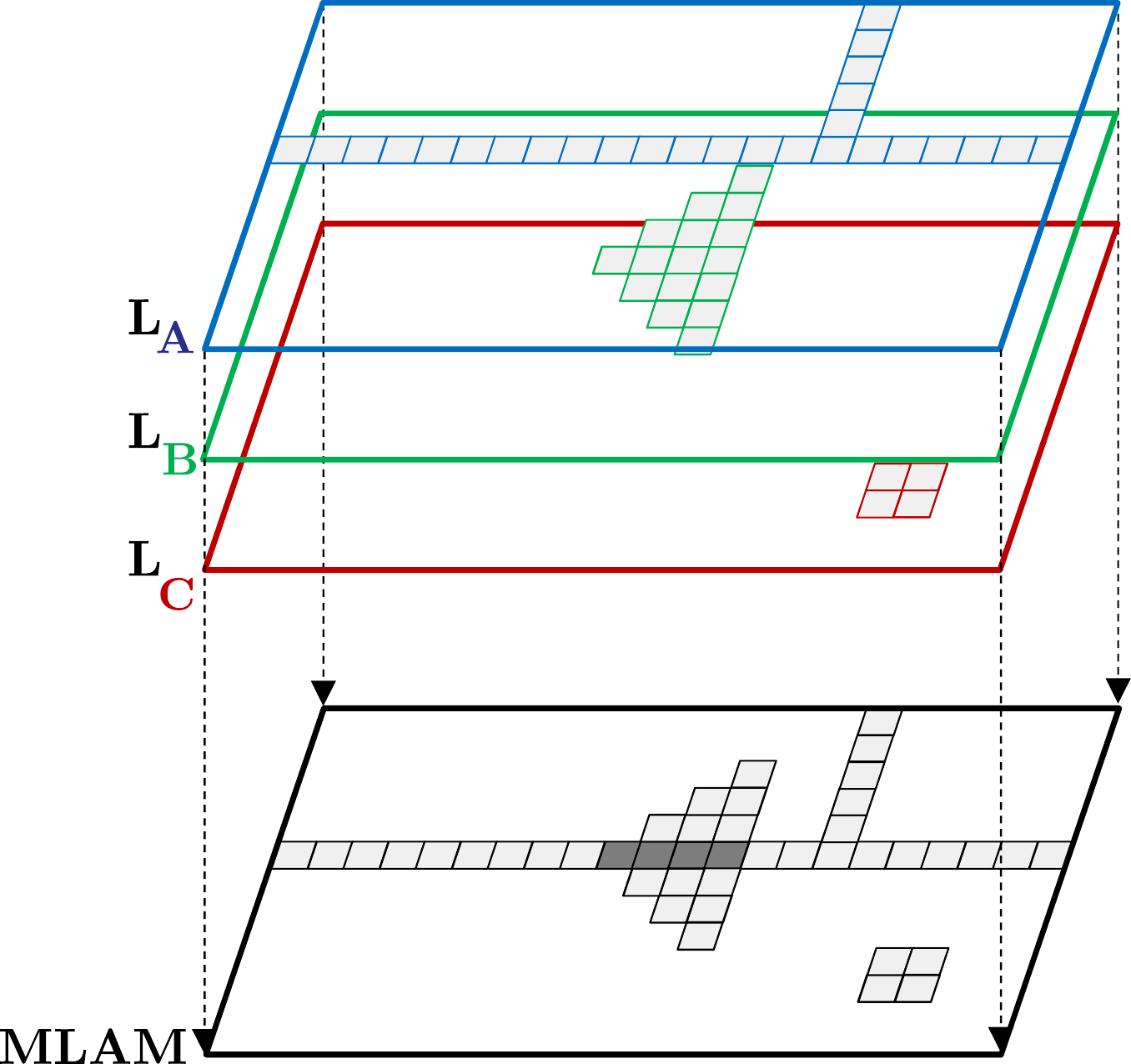}
    \caption{Example for a Cartesian Multi-Layer Attention Map (MLAM) representation constructed by the aggregation of three layers \textbf{L}. Darker cells correspond to higher required attention. Graphic taken from~\cite{henning2022}.}
    \label{fig:saep:mlam}
\end{figure}

To generate the performance requirement $p_i^\text{req}$ towards the environment perception for each region $r_i$ within the environment $\mathbb{E}$, the set of relevance defining layers $\mathbb{L}$ is introduced. 
Each layer $\mathbf{l}_k \in \mathbb{L}$ represents an independent environmental influence for the agent and assigns a performance requirement 
\begin{equation}
    p_{i,k}^{\text{req}} = \mathbf{l}_k(r_i)
\end{equation}
to any region $r_i$. 
The situational influence on the relevance of information is realized by using the current situation $\mathbf{s}$ as an activation function for a subset of relevant layer functions $\mathbb{L}_a \subseteq \mathbb{L}$.
To ensure applicability, the set of possible situations $\mathbb{S}$ corresponds to combinations of high-level situational influences, e.g., weather conditions, geographic location, or agent operation states.
The attention map, containing performance requirements
\begin{equation}
\label{eq:concept:comb}
     p_{i}^{\text{req}} = \sum_{l_k \in \mathbb{L}_a} \mathbf{l}_k(r_i)
\end{equation} 
for all regions, is then obtained by summation of the individual performance requirements from each layer. 
It is consequently labeled the \textit{multi-layer attention map} (MLAM). 
A visual example using a Cartesian grid representation for the description of the environment is shown in Fig.~\ref{fig:saep:mlam}.

\subsubsection{Processing Chain Configuration}
For the identification of the optimal environment perception software configuration, its elements are represented by a set of functional modules $m\in \mathbb{M}$. 
These modules correspond to common algorithm types for the perception of automated driving applications, e.g., object tracking, object detection, semantic segmentation, or free space detection.  
Each module $m$ provides a set of properties, comprising estimates for its resource consumption in terms of cost $c_m$, its provided benefit for the quality of environment representation w.r.t. the agents driving behavior in terms of a performance value $p_m$, and its relations $\mathcal{R}$, representing linkabilities, dependencies, or interactions to other modules or external influences. 
Besides, each module refers to a subset of regions $\mathbb{E}_m \subseteq \mathbb{E}$ where it can process data from and is either classified as a source or a non-source module.

With these properties in consideration, the optimal configuration $\mathbb{M}^\ast$ is found by searching for a module combination candidate $\mathbb{M}'$ that is valid with respect to the relations $\mathcal{R}$ of the modules and satisfies the performance requirement of the MLAM $\mathcal{P}^\text{req} =\left\{ p_i^{\text{req}}\right\}~\forall~r_i$ at lowest aggregated cost:
\begin{align}
\mathbb{M}^\ast &=  \underset {\mathbb{M}' \subseteq \mathbb{M}} {\text{arg
min}} \left( 
\sum_{m \in \mathbb{M}'} c_m  \right)\label{eq:intro:RC_min}
~\text{s.t.}~\sum_{m \in \mathbb{M}'} p_m \geq \mathcal{P}^{\text{req}}.
\end{align}
To structure the search for the optimal configuration, the relations $\mathcal{R}$ of the modules are used to generate a set of configuration trees. 
All nodes within the set of trees correspond to valid combination candidates $\mathbb{M}'$, where root nodes strictly correspond to candidates that only contain source modules. 
The coverage of the candidates, i.e., the unification of the contained modules' coverage, is evaluated against the requirements of the MLAM to invalidate insufficient trees entirely. Consequently, the search space for the optimal configuration can be reduced drastically for complex configuration trees.

Finally, the configuration of the identified, optimal subset of processing modules is realized via the underlying software architecture. ASOA~\cite{mokhtarian2020} is used within the UNICAR\emph{agil} project. Other examples of existing software architectures are~\cite{Kim2012},~\cite{Schlatow2015}, or~\cite{Thomas2014}.

\subsubsection{Attention Map Application}
In the final step of \textit{awareness processing}, the MLAM is fed into the configured software modules $m\in \mathbb{M}^\ast$. Existing methods and algorithms that consider situation-awareness, e.g., in the form of selective data processing (cf. examples from Section~\ref{sec:intro}), can be leveraged. The available system resources are hence allocated efficiently towards processing only relevant information instead of the usual uniform distribution, i.e., processing all data with equal importance.

\subsection{Application for the UNICAR\emph{agil} Vehicle}
With the summary of \textit{awareness processing} provided, we present its application to the decentralized architecture of the UNICAR\emph{agil} prototype vehicles (cf. Section~\ref{sec:intro:unicar}) in the following. 

\subsubsection{Situation Detection and Attention Map Generation}
Our project goal in UNICAR\emph{agil} is a maneuver-dependent environment perception. Thus, we define the set of situations $\mathbb{S}$ to contain all valid maneuvers, where each maneuver corresponds to a tuple between a \textit{directional maneuver} ($\text{dm}$) and a \textit{lateral maneuver} ($\text{lm}$):
\begin{equation}
    \mathbb{S} = \{ (\text{dm}, \text{lm}) \}, ~\text{with}
\end{equation}
\begin{equation}
    \begin{split}
        \text{dm} \in~ &  \{ \text{forward}, \text{backward}, \text{left}, \text{right}, \\
        & ~\text{maneuvering}, \text{standby}\},
    \end{split}
\end{equation}
\begin{equation}
    \begin{split}
        \text{lm}_y \in~ & \{\text{turn left}, \text{turn right},\\
        & ~\text{change left}, \text{change right}, \emptyset  \}.
    \end{split}
\end{equation}
No lateral maneuver is represented by $\emptyset$. The lateral maneuvers are invariant to the vehicle's directional maneuver, i.e., side definitions do not change with the driving direction.
The directional maneuvers \textit{left} and \textit{right} reflect the capability of the disruptive dynamic modules~\cite{woopen2018} of the vehicles to move sideways. During these maneuvers, no lateral maneuver will be conducted.

\begin{table}[t!]
    \centering
    \caption{Activation of attention layers per maneuver.}
    \label{tab:saep:layeractivation}
    \setlength{\tabcolsep}{2pt}
    \begin{tabular}{rccc}
    \toprule
                 & directional intention & lateral intention & maneuvering \\
    maneuver     & (\textit{DI})         & (\textit{LI})     & (\textit{M}) \\
                 \midrule
    forward      & x                     &                   &             \\
    backward     & x                     &                   &             \\
    left         & x                     &                   &             \\
    right        & x                     &                   &             \\
    maneuvering  &                       &                   & x           \\
    turn left    &                       & x                 &             \\
    turn right   &                       & x                 &             \\
    change left  &                       & x                 &             \\
    change right &                       & x                 &             \\
    standby      &                       &                   &            \\
    \bottomrule
    \end{tabular}
\end{table}
The corresponding MLAM is constructed from three attention layers, which are activated as defined in Tab.~\ref{tab:saep:layeractivation}. The corresponding layer functions $\mathbf{l}$ are represented by a direct mapping between the situation, i.e., the maneuver in our case, and the regions of the environment (cf. Fig.~\ref{fig:intro:unicar}).
The regions $r_i$ are defined as eight zones around the vehicle that are derived by considering the borders of overlapping perception areas between the sensor modules.
Each layer assigns a performance requirement of $\mathbf{l}(r_i)=1$ to mapped regions as described in the following. An overview is presented in Tab.~\ref{tab:saep:regionmapping}. 
The \textit{maneuvering} layer is active only for the identically named maneuver. It represents low-velocity, high-attention scenarios, such as roundabouts or areas shared with vulnerable road users. Consequently, it assigns relevance to all regions.
The \textit{directional intention} layer is active in all directional maneuvers. Relevance is assigned to the three regions corresponding to the maneuver.
The \textit{lateral intention} layer is active for all lateral movements, i.e., turning and lane changing. Here, the assigned relevance depends on the current directional maneuver.

\begin{table}[t!]
    \centering
    \caption{Mapping between assigned attention and maneuvers by attention layers. Abbreviations used as per Tab.~\ref{tab:saep:layeractivation}. Superscripts~\textsuperscript{f} and \textsuperscript{b} indicate the dependency on the forward and backward directional maneuver, respectively.}
    \label{tab:saep:regionmapping}
    \setlength{\tabcolsep}{5pt}
    \begin{tabular}{rcccccccc}
    \toprule
    maneuver     &    fl     &    f      &    fr     &    r      & br        &   b       &   bl      &   l \\
                 \midrule
    forward      &\textit{DI}&\textit{DI}&\textit{DI}&           &           &           &           &           \\
    backward     &           &           &           &           &\textit{DI}&\textit{DI}&\textit{DI}&           \\
    left         &\textit{DI}&           &           &           &           &           &\textit{DI}&\textit{DI}\\
    right        &           &           &\textit{DI}&\textit{DI}&\textit{DI}&           &           &           \\
    maneuvering  &\textit{M} &\textit{M} &\textit{M} &\textit{M} &\textit{M} &\textit{M} &\textit{M} &\textit{M}\\
    turn left    &\textit{LI}&           &           &           &           &           &\textit{LI}&\textit{LI}\\
    turn right   &           &           &\textit{LI}&\textit{LI}&\textit{LI}&           &           &           \\
    change left  &\textit{LI}\textsuperscript{b}&           &           &           &           &           &\textit{LI}\textsuperscript{f}&\textit{LI}\\
    change right &           &           &\textit{LI}\textsuperscript{b}&\textit{LI}&\textit{LI}\textsuperscript{f}&           &           &           \\
    standby      &           &           &           &           &           &           &           &          \\
    \bottomrule
    \end{tabular}
\end{table}

\begin{figure}[t!]
    \centering
    \subfloat[MLAM for maneuver \newline $\mathbf{s}=\{(\text{forward},\emptyset)\}$.]{
        \includegraphics[width=0.4\linewidth]{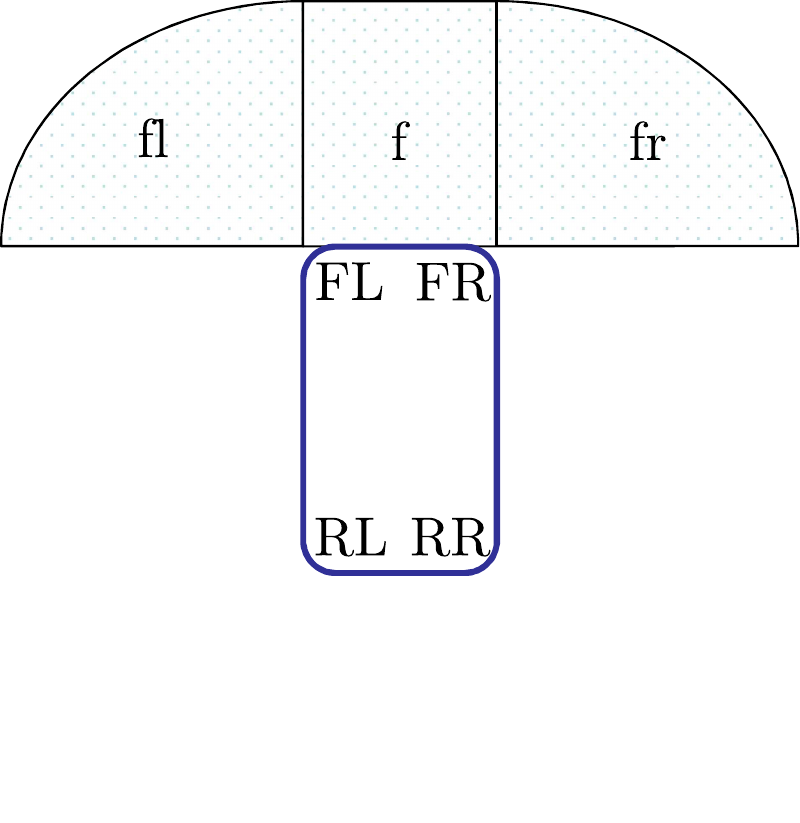}
        \label{fig:saep:mlam_example:a}}
    \hfil
    \subfloat[MLAM for maneuver \newline$\mathbf{s}=\{(\text{backward},\text{turn right})\}$.]{
        \includegraphics[width=0.4\linewidth]{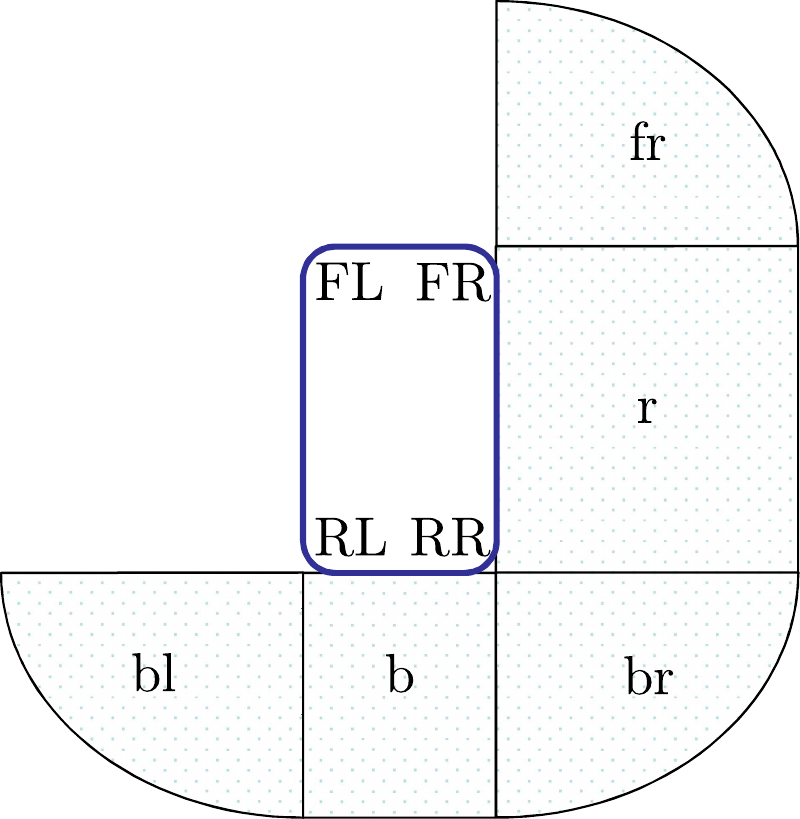}
        \label{fig:saep:mlam_example:b}}
    \caption{Visual examples for the resulting MLAM.}
    \label{fig:saep:mlam_example}
\end{figure}

In Fig.~\ref{fig:saep:mlam_example}, visual examples for the resulting attention map are presented. 
For a common $\mathbf{s}=\{(\text{forward},\emptyset)\}$ maneuver, Fig.~\ref{fig:saep:mlam_example:a} shows the vast reduction of the required perception field according to the activation of only the \textit{directional intention} layer. 
Considering a maneuver with a lateral component as in Fig.~\ref{fig:saep:mlam_example:b} for $\mathbf{s}=\{(\text{backward},\text{turn right})\}$, the resulting requirements towards the environment perception cover a more extensive set of regions, while still presenting potential for resource reduction compared to complete processing of all data in all regions.

\subsubsection{Processing Chain Configuration}

Based on the introduced decentralized functional architecture of the vehicles, the set of processing modules for environment perception is defined as the set of available sensor modules as per Fig.~\ref{fig:intro:unicar}:
\begin{align*}
   \mathbb{M} = &\{ \text{FL}, \text{FR}, \text{RL}, \text{RR}\}.
\end{align*}
Adhering to the modular concept, the properties of each sensor module are set identically, so that $\text{cost} = 1.0$, $\text{performance}=\infty$, and $\mathcal{R}=\emptyset$, i.e., no relations are to be considered. 
Besides, each module is classified as source. 
Their coverage is summarized in Tab.~\ref{tab:saep:coverage}.
Defining the performance of each module as infinite reflects the assumption that any sensor module is capable of generating a trustworthy environment model. Additional relations might be imposed, e.g., requirements for redundant coverage between sensor modules, that are out of the scope of this work.

\begin{table}[t!]
    \centering
    \caption{Module coverage of the regions (cf. Fig.~\ref{fig:intro:unicar}).}
    \label{tab:saep:coverage}
    \setlength{\tabcolsep}{15pt}
    \begin{tabular}{c|cccc}
    \toprule
       & FL & FR & RL & RR \\
       \midrule
    fl & x  & x  & x  &    \\
    f  & x  & x  &    &    \\
    fr & x  & x  &    & x  \\
    r  &    & x  &    & x  \\
    br &    & x  & x  & x  \\
    b  &    &    & x  & x  \\
    bl & x  &    & x  & x  \\
    l  & x  &    & x  &   \\
    \bottomrule 
    \end{tabular}
\end{table}

The resulting set of configuration trees contains all combinations between the sensor modules, i.e., the powerset of the set of modules $\mathscr{P}\left( \mathbb{M} \right)$, including the set $\mathbb{M}$ itself (all modules active) and the empty set $\emptyset$ (no module active). Hence, the set of configuration trees comprises $16$ root nodes, with no further leaf nodes, as no non-source modules are considered within the scope of this work.

\subsubsection{Attention Map Application}
Lastly, the attention map is provided to the configured sensor modules. 
Based on the identified relevant regions, the data processing within the functional modules of the independent sensor module environment model is restricted to a subset of the three visible \SI[mode=text]{90}{\degree}  quadrants of their \SI[mode=text]{270}{\degree} field of view.
For each sensor module, the middle quadrant coincides with the corresponding corner region, e.g., \textit{fl} for the FL module. 
Each of the two outer quadrants corresponds to a coverage of two regions, e.g., \textit{f} and \textit{fr} for the FL module's second quadrant and \textit{l} and \textit{bl} for its third.

\section{Evaluation}
Since the driverless prototype vehicles are completely built from scratch in the UNICAR\emph{agil} project, they are not yet available to test our method during run-time. However, we have a complete sensor module available in the lab~\cite{Buchholz2020} with the same hardware and software components as in the vehicles, which we use during the project development phase. Thus, it enables us to generate and process data with identical behavior and properties as during a real-world drive of such a sensor module.

\begin{figure}
    \centering
    \includegraphics[width=.6\linewidth]{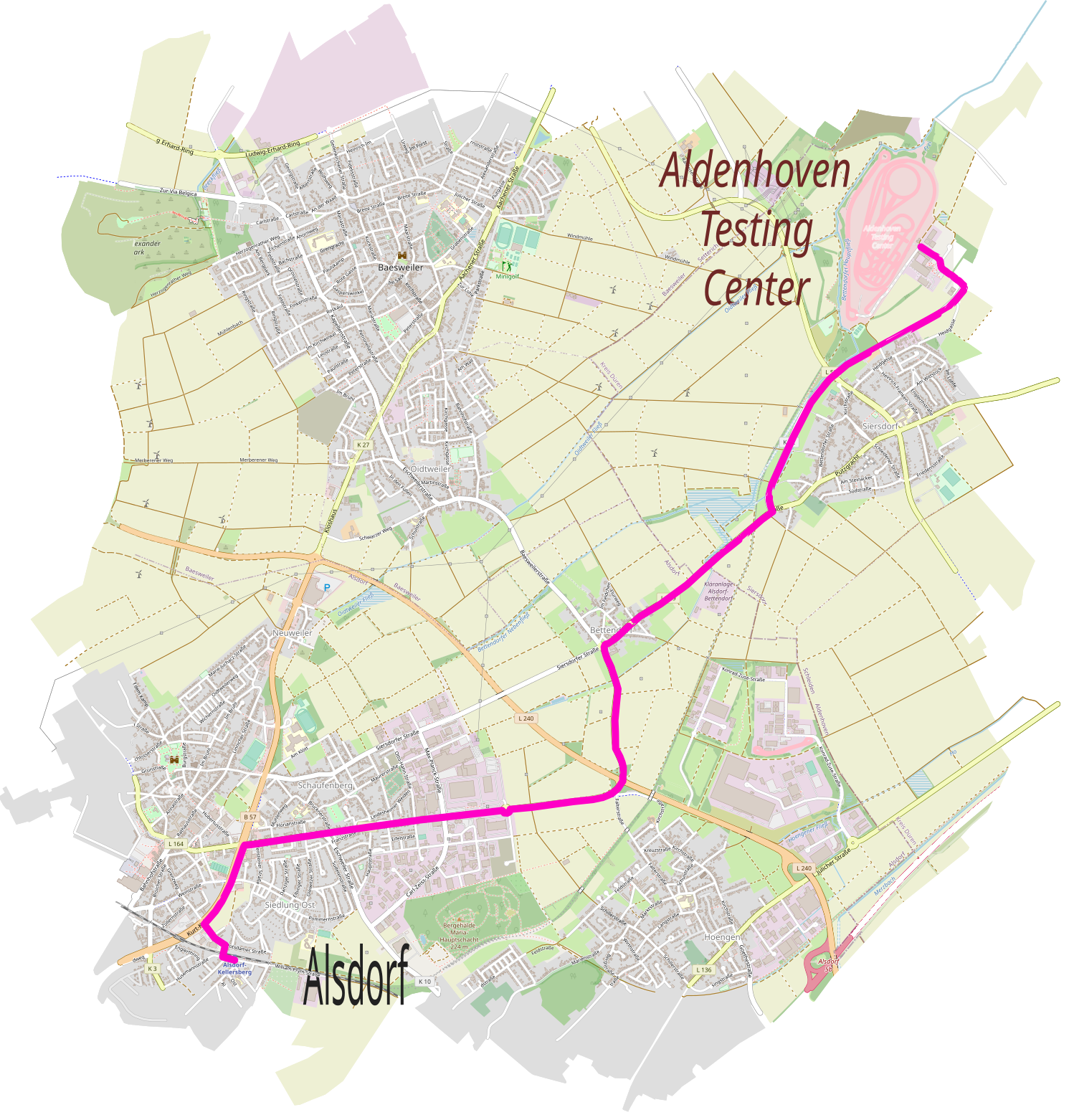}
    \caption{Outline of the hypothetical route between Alsdorf, Germany and Aldenhoven Testing Center~\cite{aldenhoven} in pink (map data from OpenStreetMap~\cite{OpenStreetMap}).
    }
    \label{fig:exp:route}
\end{figure}

To showcase the application in a realistic setup, we define a hypothetical route for the \emph{auto}SHUTTLE~\cite{woopen2018} between the Aldenhoven Testing Center~\cite{aldenhoven}, where future project presentations will be conducted, and the nearest train station in Alsdorf, Germany.
The route represents the application as an automated, close-range public transport vehicle and comprises a track length of approximately \SI[mode=text]{7.5}{\kilo\meter} at a journey-time of \SI[mode=text]{9}{\minute}~\SI[mode=text]{15}{\second}.
It includes various urban and rural driving conditions and is outlined in Fig.~\ref{fig:exp:route}.  
\begin{table}
    \centering
    \caption{Maneuver distribution during the route.}
    \begin{tabular}{rlrl}
    \toprule
    directional& \multirow{2}{*}{occurance} & lateral & \multirow{2}{*}{occurance}\\
     maneuver &  & maneuver &  \\
    \midrule
    forward              & 89.7\% & turn left        & 2.0\%  \\
    backward             & 0.0\%  & turn right       & 2.2\%  \\
    left                 & 1.5\%  & change left      & 0.7\%  \\
    right                & 1.5\%  & change right     & 1.1\%  \\
    maneuvering          & 7.3\%  & $\emptyset$      & 94.1\% \\
    standby              & 0.0\%  &                  &        \\
    \bottomrule
    \end{tabular}
    \label{tab:exp:maneuverDist}
\end{table}

The maneuver distribution during the hypothetical route is shown in Tab.~\ref{tab:exp:maneuverDist}.
The sequence of maneuvers is generated with 1Hz based on map data from OpenStreetMap~\cite{OpenStreetMap} and a simplified movement model of the vehicle. 
As expected for regular driving conditions, the \textit{forward} maneuver is dominating. The maneuvers \textit{left} and \textit{right}  are conducted only at the start and at the end of the route to represent sideways parking maneuvers.

\subsection{Module Configuration}
\begin{figure}[!t]
    \centering
    \includegraphics[width=.75\linewidth]{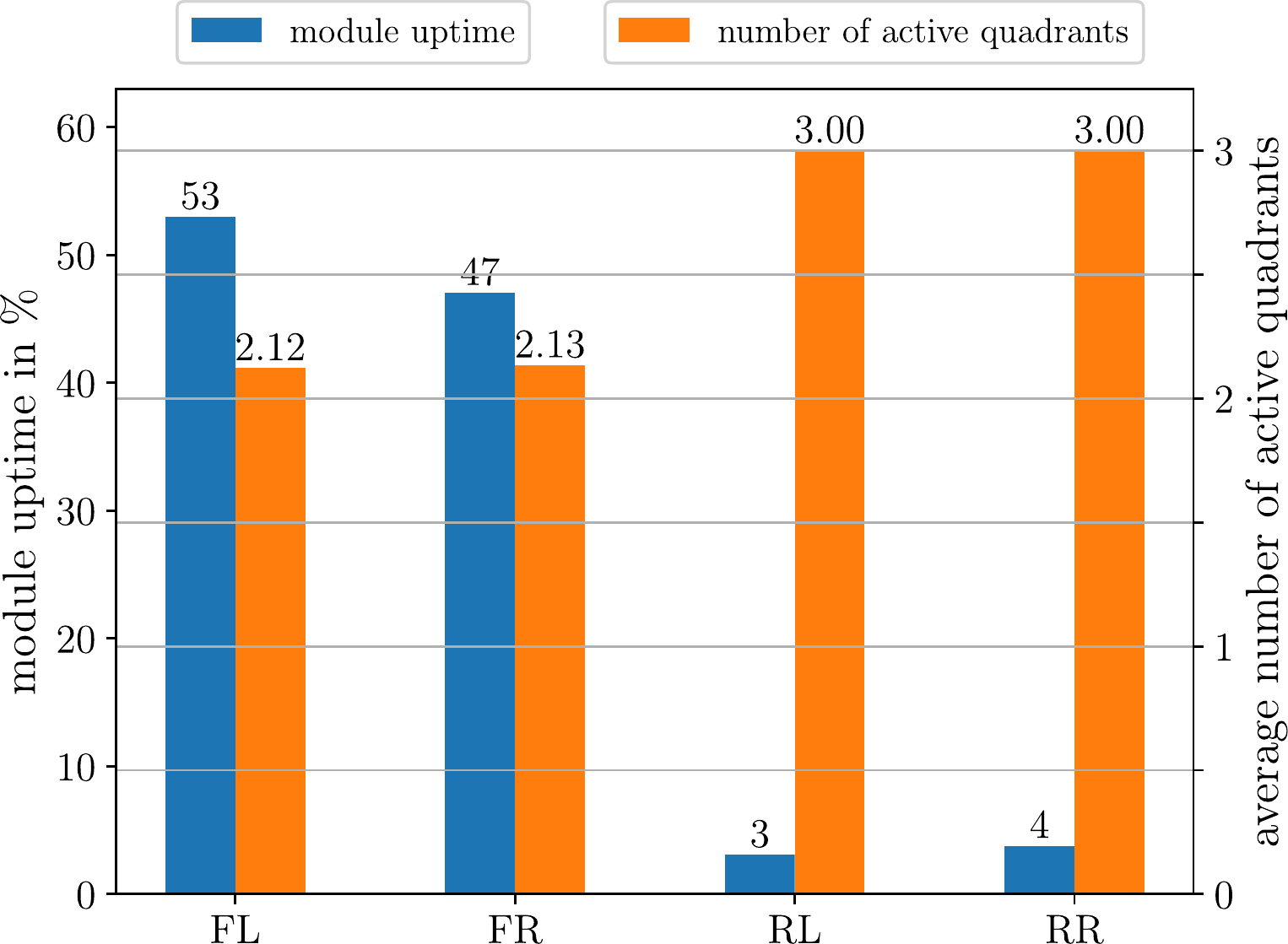}
    \caption{Summarized module configuration on the hypothetical route.}
    \label{fig:exp:uptime}
\end{figure}

We can derive the resulting module configuration using the introduced \textit{awareness processing} application along the hypothetical  route  in simulation. 
Fig.~\ref{fig:exp:uptime} shows the module uptime, referring to the number of cycles the sensor modules were configured to be active, as well as the corresponding number of active quadrants. 

The module uptime (blue) accumulates to \SI[mode=text]{100}{\percent} between the two front modules, reflecting coverage of the front area through the entire route as well as a strictly disjoint activation between these modules.
The module uptime for the front-left module exceeds the uptime of the front-right module. This behavior depends on the sequence of maneuvers, as an activation hysteresis is embedded in the simulation. Coherently, a sensor module will stay active to reduce ramp-up and ramp-down efforts and only switches to an active state if a lateral maneuver requires the opposite side to be observed.
During these lateral maneuvers, the third quadrant is active for the respective front module. 
This is reflected in the averaged number of active quadrants (orange) slightly above $2$ for both modules.
For the two rear modules, their low uptime corresponds to the low count of maneuvering situations (\SI[mode=text]{7.3}{\percent}, cf. Tab.~\ref{tab:exp:maneuverDist}). 
During these maneuvers, the rear module on the opposite side of the currently active front module is configured to ensure a \SI[mode=text]{360}{\degree} coverage of the environment, as required by the MLAM parameterization.
The averaged number of active quadrants verifies the functionality, as for every active cycle of a rear-side module, all quadrants are active, and the average equals exactly $3.00$.

\subsection{System Power Consumption}
To provide quantitative results of the energy reduction potential of \textit{awareness processing} using the introduced MLAM application, i.e., restricting the processing of sensor data to active quadrants within active sensor modules, we leverage measurements of the power consumption of the available sensor module in the lab.
For the scope of our experiments, we have generated a data set within the vicinity of the Ulm University, Germany, of roughly the duration of the hypothetical route. To ensure reproducibility, the data evaluation is conducted in a post-processing manner. The generated dataset is processed fully per configuration, and the power consumption is averaged from 100 samples per second.

\begin{figure}[!t]
    \centering
    \includegraphics[width=.75\linewidth]{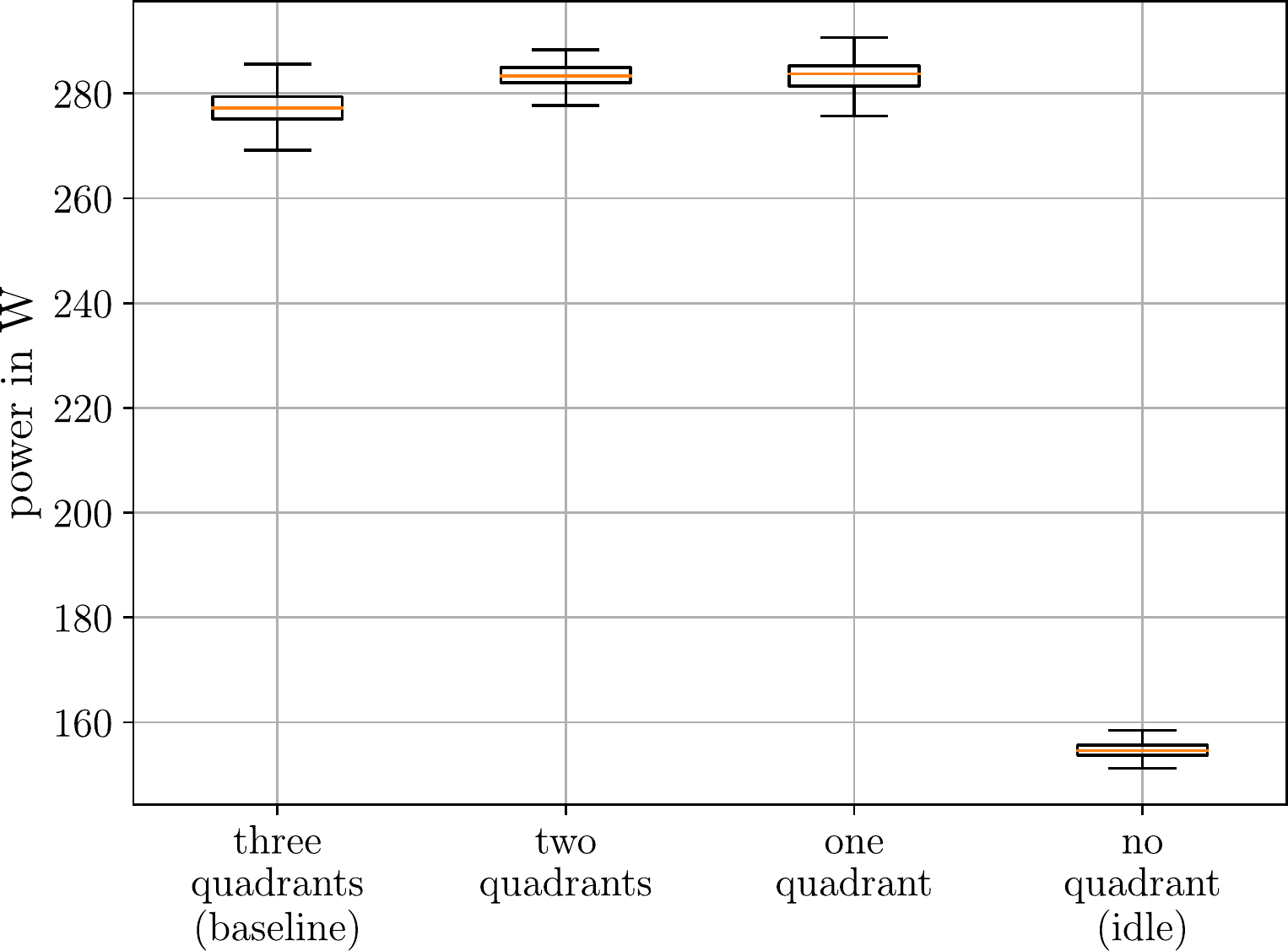}
    \caption{Power consumption per number of active quadrants.}
    \label{fig:exp:segmentPower}
\end{figure}
Fig.~\ref{fig:exp:segmentPower} provides the resulting distribution of a sensor module's power consumption per number of active quadrants as boxplots. 
Contradictory to the initial assumption of resource reduction, the figure shows that by reducing the processed data to one or two quadrants, the median consumption increases by approximately \SI[mode=text]{2.3}{\percent} compared to the baseline with all three quadrants being active. The spread of power consumption remains largely unaffected in both cases.
By reducing the processing to zero processing quadrants, i.e., configuring the module into a standby state, the median consumption is reduced by approximately \SI[mode=text]{44}{\percent}, complying with the assumptions.
Hence, limiting data processing to active quadrants is ineffective in reducing the power consumption of a sensor module in its current configuration.

Considering the software modules for environment perception (cf. ~\cite{Buchholz2020}), we conclude that the used deep learning architectures of the object detection modules used in this work are not suitable for an effective MLAM application. For example, the underlying PointNet~\cite{pointnet2017} and PointPillar~\cite{pointpillars2019} architectures used for lidar and radar data use a fixed number of inputs, independent of the field-of-view currently processed, leading to a fixed number of internal calculations. The MLAM application induces additional processing effort to convert the reduced size of input data to the expected size, e.g., by padding and copying operations. Since we have already shown in our previous work (cf. ~\cite{henning2022}) that hardware load can be reduced by MLAM application for other software components, we conclude in compliance with Bajscy et al.~\cite{Bajcsy2018} that research within the field of \textit{awareness processing} should be strengthened. To fully leverage a distinction between relevant and non-relevant information, either employed software modules must be designed scalable, or vehicle architectures need to contain configurable module options that are specifically designed for the processing of data regions instead of full sensor ranges.

Nonetheless, even from the module configuration alone, a significant reduction in power consumption could be achieved.
To obtain an estimate of the reduction in power consumption for the combined perception architecture, we aggregate the results presented in Fig.~\ref{fig:exp:uptime} and Fig.~\ref{fig:exp:segmentPower} for the hypothetical route. 
Traversing the route once, the estimated power consumption drops by \SI[mode=text]{31.9}{\percent} from \SI[mode=text]{0.171}{\kilo\watt\hour} for the baseline (all four sensor modules fully active at all times) to \SI[mode=text]{0.116}{\kilo\watt\hour} with \textit{awareness processing}.
Further, we scale the hypothetical application to an entire working day of 10 hours, assuming boarding times of \SI[mode=text]{5}{\minute} between every transfer, where all sensor modules are in a standby state.  For the baseline, the resulting power consumption yields \SI[mode=text]{11.06}{\kilo\watt\hour}, while the power consumption with \textit{awareness processing} is only \SI[mode=text]{7.05}{\kilo\watt\hour}. That means that even without the MLAM application being effective in this example, \textit{awareness processing} allows for an estimated reduction in resource consumption of the system's environment perception of impressive \SI[mode=text]{36.2}{\percent}.

\section{Conclusions}
In this work, we have shown the applicability of our previously published concept for \textit{awareness processing}~\cite{henning2022} towards a decentralized software and processing architecture and disruptive vehicle  dynamics, including, e.g., sideways movement. Representing the future use of one of the UNICAR\emph{agil} prototype vehicles~\cite{Woopen2020} as a short-range shuttle service, we have generated a hypothetical route for which the daily power consumption of the systems environment perception is estimated to reduce by \SI[mode=text]{36.2}{\percent} by application of \textit{awareness processing}.
While this reduction is very promising, we found out that the current sensor modules' software components  do not allow for energy consumption reduction by regional separation of data to be processed using our MLAM concept. 

We conclude that the potential of situation-awareness can only be leveraged fully if automated vehicles are designed to incorporate regionally separated data processing. 
Thus, with this work, we aspire to accelerate the inclusion of situation-awareness within the field of perception in automated driving applications.
For the continued development within the UNICAR\emph{agil} project, we will similarly pursue the outlined challenges.

\addtolength{\textheight}{-12cm}   




\end{document}